\documentclass[11pt]{article}

\usepackage{xr}
\usepackage{xspace}

\usepackage[utf8]{inputenc} 
\usepackage[T1]{fontenc}    
\usepackage{hyperref}       
\usepackage{url}            
\usepackage{booktabs}       
\usepackage{amsfonts}       
\usepackage{nicefrac}       
\usepackage{microtype}      
\usepackage{graphicx}
\usepackage{subcaption}
\usepackage{color, comment}
\usepackage{bm, stmaryrd}
\usepackage{fullpage}
\usepackage{algorithm}
\usepackage{algorithmic}
\usepackage{footnote}
\usepackage{tablefootnote}
\usepackage{amsmath, amsthm, amssymb}

\newtheorem{theorem}{Theorem}
\newtheorem{corollary}{Corollary}
\newtheorem{assumption}{Assumption}
\newtheorem{lemma}{Lemma}

\newtheorem{remark}{Remark}

\newcommand{\bR}{\mathbb{R}} 
\renewcommand{\dim}{p}

\newcommand{\cI}{\mathcal{I}}

\newcommand{\cC}{\mathcal{C}}

\newcommand{\tC}{\widetilde{C}}

\newcommand{\cR}{\mathcal{R}}
\newcommand{\cM}{\mathcal{M}}
\newcommand{\bE}{\mathbb{E}}

\newcommand{\tr}{\text{trace}}
\newcommand{\cO}{\mathcal{O}}

\newcommand{\bfone}{\textbf{1}}
\newcommand{\xhat}{\hat{X}}
\newcommand{\cl}{r}
\newcommand{\xh}{\hat{X}}

\newcommand{\innerprod}[2]{\langle {#1}, {#2}\rangle}

\newcommand{\bas}[1]{\begin{align*}#1\end{align*}}
\newcommand{\ba}[1]{\begin{align}#1\end{align}}
\newcommand{\sgnorm}{b}
\newcommand{\kp}{\tilde{K}}
\newcommand{\km}{$k$-means\xspace}
\newcommand{\ksvd}{K-SVD\xspace}
\newcommand{\kpca}{K-PCA\xspace}

\newcommand{\gm}{\gamma_{\min}}

\newcommand{\vertiii}[1]{{\left\vert\kern-0.25ex\left\vert\kern-0.25ex\left\vert #1 
    \right\vert\kern-0.25ex\right\vert\kern-0.25ex\right\vert}}

\title{On Robustness of Kernel Clustering}

\author{
  Bowei Yan\\
  University of Texas at Austin\\
  \and
  Purnamrita Sarkar \\
  University of Texas at Austin\\
}
\date{}
\begin{document}

\maketitle

\begin{abstract}
Clustering is an important unsupervised learning problem in machine learning and statistics. Among many existing algorithms, kernel \km has drawn much research attention due to its ability to find non-linear cluster boundaries and its inherent simplicity. There are two main approaches for kernel k-means: SVD of the kernel matrix and convex relaxations. Despite the attention kernel clustering has received both from theoretical and applied quarters, not much is known about robustness of the methods. In this paper we first introduce a semidefinite programming relaxation for the kernel clustering problem, then prove that under a suitable model specification, both K-SVD and SDP approaches are consistent in the limit, albeit SDP is strongly consistent, i.e. achieves exact recovery, whereas K-SVD is weakly consistent, i.e. the fraction of misclassified nodes vanish. Also the error bounds suggest that SDP is more resilient towards outliers, which we also demonstrate with experiments.
\end{abstract}

\section{Introduction}
\label{sec:intro}
Clustering is an important problem which is prevalent in a variety of real world problems. One of the first and widely applied clustering algorithms is \km, which was named by James MacQueen \cite{MacQueen}, but was proposed by Hugo Steinhaus~\cite{steinhaus} even before. Despite being half a century old, \km has been widely used and analyzed under various settings.


One major drawback of \km is its incapability to separate clusters that are non-linearly separated. This can be alleviated by mapping the data to a high dimensional feature space and do clustering on top of the feature space \cite{scholkopf1998nonlinear, dhillon2004kernel, kim2005evaluation}, which is generally called kernel-based methods. For instance, the widely-used spectral clustering \cite{shi2000normalized, ng2002spectral} is an algorithm to calculate top eigenvectors of a kernel matrix of affinities, followed by a \km on the top $\cl$ eigenvectors. The consistency of spectral clustering is analyzed by \cite{von2008consistency}. \cite{dhillon2004kernel} shows that spectral clustering is essentially equivalent to a weighted version of kernel \km.

The performance guarantee for clustering is often studied under distributional assumptions; usually a mixture model with well-separated centers suffices to show consistency. \cite{dasgupta2007probabilistic} uses a Gaussian mixture model, and proposes a variant of EM algorithm that provably recovers the center of each Gaussian when the minimum distance between clusters is greater than some multiple of the square root of dimension. \cite{awasthi2012improved} works with a projection based algorithm and shows the separation needs to be greater than the operator norm and the Frobenius norm of difference between data matrix and its corresponding center matrix, up to a constant.

Another popular technique is based on semidefinite relaxations. For example \cite{kulis2007fast, peng2007approximating} propose SDP relaxations for \km typed clustering. 
In a very recent work, \cite{mixon2016clustering} shows the effectiveness of SDP relaxation with \km clustering for subgaussian mixtures, provided the minimum distance between centers is greater than the variance of the sub-gaussian times the square of the number of clusters $\cl$.

On a related note, SDP relaxations have been shown to be consistent for community detection in networks \cite{amini2014semidefinite, 
cai2015robust}. In particular, \cite{cai2015robust} consider ``inlier'' (these are generated from the underlying clustering model, to be specific, a blockmodel) and ``outlier'' nodes. The authors show that SDP is weakly consistent in terms of clustering the inlier nodes as long as the number of outliers $m$ is a vanishing fraction of the number of nodes.

In contrast, among the numerous work on clustering, not much focus has been on robustness of different kernel \km algorithms in presence of arbitrary outliers. \cite{yang2004similarity} illustrates the robustness of Gaussian kernel based clustering, where no explicit upper bound is given. \cite{debruyne2010detecting} detects the influential points in kernel PCA by looking at an influence function. In data mining community, many find clustering can be used to detect outliers, with often heuristic but effective procedures \cite{pamula2011outlier,duan2009cluster}. On the other hand, kernel based methods have been shown to be robust for many machine learning tasks. For supervised learning, \cite{xu2006robust} shows the robustness of SVM by introducing an outlier indicator and relaxing the problem to a SDP. \cite{de2009robustness, debruyne2008model, christmann2007consistency} develop the robustness for kernel regression.  For unsupervised learning, \cite{kim2012robust} proposes a robust kernel density estimation.

In this paper we ask the question: how robust are SVD type algorithms and SDP relaxations when outliers are present. In the process we also present results which compare these two methods. To be specific, we show that without outliers, SVD is weakly consistent, i.e. the \textit{fraction} of misclassified nodes vanishes with high probability, whereas SDP is strongly consistent, i.e. the \textit{number} of misclassified nodes vanishes with high probability. We also prove that both methods are robust to arbitrary outliers as long as the number of outliers is growing at a slower rate than the number of nodes. Surprisingly our results also indicate that SDP relaxations are more resilient to outliers than K-SVD methods. The paper is organized as follows. 
In Section \ref{sec:setup} we set up the problem and the data generating model. We present the main results in Section \ref{sec:mainres}. Proof sketch and more technical details are introduced in Section \ref{sec:analysis}. Numerical experiments in Section \ref{sec:exp} illustrate and support our theoretical analysis. 

\section{Problem Setup} 
\label{sec:setup}

We denote by $Y=[Y_1,\cdots,Y_{n}]^T$ the $n\times p$ data matrix. Among the $n$ observations, $m$ outliers are distributed arbitrarily, and $n-m$ inliers form $\cl$ equal-sized clusters, denoted by $C_1,\cdots, C_\cl$. Let us denote the index set of inliers by $\cI$ and index set of outliers by $\cO$, $\cI\cup\cO=[n]$. Also denote by $\cR=\{(i,j): i\in \cO \text{ or }j\in \cO\}$.

The problem is to recover the true and unknown data partition given by a membership matrix $Z=\{0,1\}^{n \times \cl}$, where $Z_{ik}=1$ if $i$ belongs to the $k$-th cluster and 0 otherwise. For convenience we assume the outliers are also arbitrarily equally assigned to $\cl$ clusters, so that each extended cluster, denoted by $\tC_i,i\in [r]$ has exactly $n/\cl$ points.
A ground truth clustering matrix $X_0\in \bR^{ n \times n}$ can be achieved by $X_0=ZZ^T$. It can be seen that $X_0(i,j)=\begin{cases} 1 & \text{ if } i,j\text{ belong to the same cluster;} \\ 0  & \mbox{ otherwise}. \end{cases} $

For the inliers, we assume the following mixture distribution model.
\begin{equation}
\begin{split}
\mbox{Conditioned on $Z_{ia}=1$, } \ \ \  &Y_i = \mu_{a}+  \frac{W_i}{\sqrt{\dim}},\ \bE[W_i]=0,\ Cov[W_i] = \sigma_a^2I_p, \\
&  \mbox{$W_i$ are independent sub-gaussian random vectors.}  \nonumber
\label{eq:data-model}
\end{split}
\end{equation}

We treat $Y$ as a low dimensional signal hidden in high dimensional noise. More concretely $\mu_a$ is sparse and $\|\mu_a\|_0$ does not depend on $n$ or $\dim$; as $n \rightarrow \infty$, $\dim\to \infty$. $W_i$'s for $i\in [n]$ are independent. For simplicity, we assume the noise is isotropic and the covariance only depends on the cluster. The sub-gaussian assumption is non-parametric and includes most of the commonly used distribution such as Gaussian and bounded distributions. We include some background materials on sub-gaussian random variables in Appendix \ref{sec:sub-gaussian}.
This general setting for inliers is common and also motivated by many practical problems where the data lies on a low dimensional manifold, but is obscured by high-dimensional noise~\cite{el2010information}.

We use the kernel matrix based on Euclidean distances between covariates. Our analysis can be extended to inner product kernels as well.  
From now onwards, we will assume that the function generating the kernel is bounded and Lipschitz.

\begin{assumption}
For $n$ observations $Y_1,\cdots, Y_n$, the kernel matrix (sometimes also called Gram matrix) $K$ is induced by $K(i,j)=f(\|Y_i-Y_j\|_2^2)$, where $f$ satisfies:
\begin{align*}
|f(x)|\le 1,\ \forall x \ \text{ and }\ \exists C_0>0, s.t. \ \sup_{x,y}|f(x)-f(y)|\le C_0|x-y|. 
\end{align*}
\label{ass:kernel}
\end{assumption}
A widely used example that satisfies the above condition is the Gaussian kernel. For simplicity, we will without loss of generality assume $K(x,y)=f(\|x-y\|^2)=\exp(-\eta \|x-y\|^2)$.

For the asymptotic analysis, we use the following standard notations for approximated rate of convergence. $T(n)$ is $O(f(n))$ iff for some constant $c$ and $n_0$, $T(n)\le cf(n)$ for all $n\ge n_0$; $T(n)$ is $\Omega(f(n))$ if for some constant $c$ and $n_0$, $T(n)\ge cf(n)$ for all $n\ge n_0$; $T(n)$ is $\Theta(f(n))$ if $T(n)$ is $O(f(n))$ and $\Omega(f(n))$; $T(n)$ is $o(f(n))$ if $T(n)$ is $O(f(n))$ but not $\Omega(f(n))$. $T(n)$ is $o_P(f(n))$ ( or $O_P(f(n))$) if it is $o(f(n))$ ( or $O(f(n))$) with high probability.

Several matrix norms are considered in this manuscript. Assume $M\in \bR^{n\times n}$, the $\ell_1$ and $\ell_\infty$ norm are defined the same as the vector $\ell_1$ and $\ell_\infty$ norm $ \|M\|_1=\sum_{ij}|M_{ij}|,\, \|M\|_\infty=\max_{i,j} |M_{ij}| $. For two matrices $M, Q \in \bR^{m\times n}$, their inner product is $\innerprod{M}{Q}=\tr(M^TQ)$. 
Let the eigenvalues of $M$ be denoted by $\lambda_1\geq \dots\geq \lambda_n$.  The operator norm $\|M\|$ is simply the largest eigenvalue of $M$, i.e. $\lambda_1$. For a symmetric matrix, it is the magnitude of the largest eigenvalue. The nuclear norm is $\|M\|_*=\sum_{i=1}^n \sigma_i$. Throughout the manuscript, we use ${\bf{1}}_n$ to represent the all one $n \times 1$ vector and $E_n,E_{n,k}$ to represent the all one matrix with size $n\times n$ and $n \times k$. The subscript will be dropped when it is clear from context.

\subsection{Two kernel clustering algorithms}
Kernel clustering algorithms can be broadly divided into two categories; one is based on semidefinite relaxation of the \km objective function and the other is eigen-decomposition based, like kernel PCA, spectral clustering, etc. In this section we describe these two settings. 
\paragraph{SDP relaxation for kernel clustering}
It is well known \cite{dhillon2004kernel} that kernel \km could be achieved by maximizing $\tr(Z^TKZ)$ where $Z$ is the $n\times r$ matrix of cluster memberships. However due to the non-convexity of the constraints, the problem is NP-hard. Thus lots of convex relaxations are proposed in literature. In this paper, we propose the following semidefinite programming relaxation. The same relaxation has been used in stochastic block models \cite{amini2014semidefinite} but to the best of our knowledge, this is the first time it is used to solve kernel clustering problems and shown to be consistent.

\begin{align}
\max_{X} &\ \tr(KX) \label{eq:sdp1} \tag{SDP-1} \\
\text{s.t.,} &\  X\succeq 0, X\ge 0, \ X{\bf{1}}=\frac{n}{\cl} {\bf{1}},\ \text{diag}(X)={\bf{1}} \nonumber
\end{align}
The clustering procedure is listed in Algorithm \ref{alg:sdp}.
\begin{algorithm}[h!]
\caption{SDP relaxation for kernel clustering}
\label{alg:sdp}
\begin{algorithmic}[1]
\REQUIRE Observations $Y_1,\cdots,Y_n$, kernel function $f$.
\STATE Compute kernel matrix $K$ where $K(i,j)=f(\|Y_j-Y_j\|_2^2)$;
\STATE Solve SDP-1 and let $\xhat$ be the optimal solution;
\STATE Do \km on the $\cl$ leading eigenvectors $U$ of $\hat{X}$.
\end{algorithmic}
\end{algorithm}

\paragraph{Kernel singular value decomposition}
Kernel singular value decomposition (\ksvd) is a spectral based clustering approach. One first does SVD on the kernel matrix, then applies \km on first $r$ eigenvectors. Different variants include \kpca \cite{scholkopf1998nonlinear}, which uses singular vectors of centered kernel matrix and spectral clustering \cite{ng2002spectral}, which uses singular vectors of normalized graph laplacian of the kernel matrix.
The detailed algorithm is shown in Algorithm \ref{alg:ksvd}. 
\begin{algorithm}[h!]
\caption{\ksvd (\kpca, spectral clustering)}
\label{alg:ksvd}
\begin{algorithmic}[1]
\REQUIRE Observations $Y_1,\cdots,Y_n$, kernel function $f$.
\STATE Compute kernel matrix $K$ where $K(i,j)=f(\|Y_j-Y_j\|_2^2)$;
\IF{\kpca}
\STATE $K \leftarrow K-K11^T/n-11^TK/n+11^TK11^T/n^2$;
\ELSIF{spectral clustering}
\STATE $K\leftarrow D^{-1/2}KD^{-1/2}$ where $D = \text{diag}(K1_n)$;
\ENDIF
\STATE Do \km on the $\cl$ leading singular vectors $V$ of $K$.
\end{algorithmic}
\end{algorithm}

\section{Main results}
\label{sec:mainres}
In this section we summarize our main results. In this paper we analyze SDP relaxation of kernel \km and \ksvd type methods. Our main contribution is two-fold. First, we show that SDP relaxation produces strongly consistent results, i.e. the number of misclustered nodes goes to zero with high probability when there are no outliers, without rounding. On the other hand, \ksvd is weakly consistent, i.e. fraction of misclassified nodes goes to zero when there are no outliers. 

In presence of outliers, we see an interesting dichotomy in the behaviors of these two methods. We present upper bounds on the number of outliers, such that the output does not contain clusters that are purely consist of outliers. We see that SDP can tolerate more outliers than \ksvd. When the number of outliers is controlled, both methods can be proven to be weakly consistent in terms of misclassification error. However, SDP is more resilient to the effect of outliers than \ksvd, if the number of clusters grows or if the separation between the cluster means decays.

Our analysis is organized as follows. First we present a result on the concentration of kernel matrix around its population counterpart. The population kernel matrix for inliers is blockwise constant with $\cl$ blocks (except the diagonal, which is one).
Next we prove that as $n$ increases, the optima $\hat{X}$ of \eqref{eq:sdp1} converges strongly to $X_0$, when there are no outliers and weakly if the number of outliers grows slowly with $n$. 
Then we show the eigenvectors of $\xh$ and $K$ are close to those of their reference matrices, which are piecewise constant aligned with the true clustering structure. We further analyze the \km step with the eigenvectors as input, to present the conditions on the number of outliers, under which the inliers are clustered into exactly $r$ clusters.
Finally we show the mis-clustering error of the clustering returned by Algorithm~\ref{alg:sdp} goes to zero with probability tending to one as $n\rightarrow\infty$ when there are no outliers; and when the number of outliers is growing slowly with $n$, the fraction of mis-clustered nodes from algorithms~\ref{alg:sdp} and~\ref{alg:ksvd} converges to zero.

We will start with the concentration of the kernel matrix.  We show that under our data model Eq.~\eqref{eq:data-model} the empirical kernel matrix with the Gaussian kernel restricted on inliers concentrates around a "population" matrix $\kp^{\cI\times\cI}$, and the $\ell_\infty$ norm of $K_f^{\cI\times \cI}-\kp_f^{\cI\times \cI}$ goes to zero at the rate of $O(\sqrt{\frac{\log \dim}{\dim}})$. We extend the $\kp$ on the outlier points to be consistent with $Z$.

\begin{theorem}
Let $d_{k\ell}=\|\mu_k-\mu_\ell\|$, and $Z_i=k, Z_j=\ell$, define
 \begin{align}\label{eq:kpop} \kp_f(i,j)=\left\{\begin{array}{ll}
f(d_{k\ell}^2+\sigma_k^2+\sigma_\ell^2) & \quad \text{ if }i\ne j,\\
f(0) & \quad \text{ if }i= j.\end{array}\right. .\end{align}
Then  there exists constant $\rho>0$, such that with probability at least $1-n^2\dim^{-\rho c^2}$,
\[\sup_{i,j\in \cI} |K_{ij}-\tilde{K}_{ij}|\le c\sqrt{\frac{\log \dim}{\dim}}.\]

\label{thm:kernel_tail}
\end{theorem}
{\remark{Setting $c=\sqrt{\frac{3\log n}{p\log p}}$, there exists constant $\rho>0$, such that  $P\left(\|K-\kp\|_\infty \ge \sqrt{\frac{3\log n}{\rho\dim}}\right) \le \frac{1}{n}.$ The error probability goes to zero for a suitably chosen constant as long as $p$ is growing faster than $\log n$.}}

While our analysis is inspired by~\cite{el2010information}, there are two main differences. First we have a mixture model where the population kernel is blockwise constant. Second, we obtain $\sqrt{\frac{\log\dim}{\dim}}$ \textit{rates} of convergence by carefully bounding the tail probabilities.  In order to attain this we further assume that the noise is sub-gaussian and isotropic. From now on we will drop the subscript $f$ and refer to the kernel matrix as $K$. 


By definition, $\kp$ is blockwise constant with $\cl$ unique rows (except the diagonal elements which are ones). An important property of $\kp$ is that $\lambda_\cl-\lambda_{\cl+1}$ (where $\lambda_i$ is the $i^{th}$ largest eigenvalue of $\kp$) will be $\Omega(n\lambda_{\min}(B)/\cl)$. $B$ is the $\cl \times \cl$ Gaussian kernel matrix generated by the centers.

\begin{lemma}
If the scale parameter in Gaussian kernel is non-zero, and none of the clusters shares a same center, let $B$ be the $\cl \times \cl$ matrix where $B_{k\ell}=f(\|\mu_k-\mu_\ell\|)$,
then
\[ \lambda_\cl(\kp)-\lambda_{\cl+1}(\kp) \ge  \frac{n}{\cl}\lambda_{\min}(B)\cdot \min_k\left(f(\sigma_k^2)\right)^2 - 2\max_k(1-f(2\sigma_k^2))=\Omega(n\lambda_{\min}(B)/\cl)\]
\label{lem:eigengap_k}
\end{lemma}

Now we present our result on the consistency of \eqref{eq:sdp1}. To this end, we will upper bound $\|\hat{X}-X_0\|_1$, where $\hat{X}$ is the optima returned by \eqref{eq:sdp1} and $X_0$ is the true clustering matrix.  We first present a lemma, which is crucial to the proof of the theorem. Before doing this, we define 
\begin{align}
\gamma_{k\ell}:= f(2\sigma_k^2)-f(d_{k\ell}^2+\sigma_k^2+\sigma_\ell^2); \qquad \quad \gm:= \min_{\ell\neq k} \ \gamma_{k\ell}
\end{align}
The first quantity $\gamma_{k\ell}$ measures separation between the two clusters $k$ and $\ell$. The second quantity measures the smallest separation possible. We will assume that $\gamma_{min}$ is positive. This is very similar to the analysis in asymptotic network analysis where strong assortativity is often assumed. Our results show that the consistency of clustering deteriorates as $\gamma_{min}$ decreases.
\begin{lemma}
Let $\xhat$ be the solution to \eqref{eq:sdp1}, then
\begin{align}
\|X_0-\xh\|_1   \le\frac{ 2\innerprod{K-\kp}{\xhat-X_0} }{\gm} \label{eq:ub}
\end{align}
\label{lem:ell1_dev}
\end{lemma}

Combining the above with the concentration of $K$ from  Theorem~\ref{thm:kernel_tail} we have the following result:
\begin{theorem}
When $d_{k\ell}^2>|\sigma_k^2-\sigma_\ell^2|, \forall k\ne \ell$, and $\gamma_{\min}=\Omega \left(\sqrt{\frac{\log \dim}{\dim}}\right)$ then for some absolute constant $c>0$, $\|X_0-\xhat\|_1\le \max\left\{o_P(1),o_P\left( \frac{mn}{\cl\gm}\right)\right\}$.
\label{th:cst_sdp1}
\end{theorem}

\begin{remark}
When there's no outlier in the data, i.e., $m=0$, $\xhat=X_0$ with high probability and SDP-1 is strongly consistent without rounding. When $m>0$, the right hand side of the inequality is dominated by $mn/\cl$. Note that $\|X_0\|_1=\frac{n^2}{\cl}$, therefore after suitable normalization, the error rate goes to zero with rate $O(m/(n\gamma_{min}))$ when $n\to \infty$.
\end{remark}

Although $\hat{X}$ is consistent to the ground truth clustering matrix, in practice one often wants to get the labeling in addition to the $X_0$. Therefore it is usually needed to carry out the last eigen-decomposition step in Algorithm~\ref{alg:sdp}. Since $X_0$ is the clustering matrix, its principal eigenvectors are blockwise constant. In order to show small mis-clustering error one needs to show that the eigenvectors of $\hat{X}$ are converging (modulo a rotation) to those of $X_0$. This is achieved by a careful application of Davis-Kahan theorem, a detailed discussion of which is deferred to the analysis in Section~\ref{sec:analysis}. 

The Davis-Kahan theorem lets one bound the deviation of the $\cl$ principal eigenvectors $\hat{U}$ of a Hermitian matrix $\hat{M}$, from the $\cl$ principal eigenvectors $U$  of  $M$  as : $\|\hat{U}-UO\|_F\leq 2^{3/2}\|M-\hat{M}\|_F/(\lambda_\cl-\lambda_{\cl+1})$ \cite{yu2015useful}, where $\lambda_\cl$ is the $\cl^{th}$ largest eigenvalue of $M$ and $O$ is the optimal rotation matrix. For a complete statement of the theorem see Appendix \ref{sec:proof_mis}. 

Applying the result to $X_0$ and $\kp$ provides us with two different upper bounds on the distance between leading eigenvectors. We will see in Theorem \ref{th:mis_sdp} that the eigengap derived by two algorithms differ, which results in different tolerance for number of outliers and upper bounds for number of misclustered nodes.
Since the Davis-Kahan bounds are tight up-to a constant~\cite{yu2015useful}, despite being upper bounds, this indicates that algorithm~\ref{alg:sdp} is less sensitive to the separation between cluster means than Algorithm~\ref{alg:ksvd}.

To analyze the \km step with eigenvectors being the input, note that \km assigns each row of $\hat{U}$ (input eigenvectors of $K$ or $\xh$) to one of $\cl$ clusters. One of the common hurdles for clustering with outliers is that one mistakenly takes the outliers as separate clusters and miss out or merge the inlier clusters in the \km step. Let $c_1\cdots, c_n\in \bR^\cl$ be defined such that $c_i$ is the centroid corresponding to the $i^{th}$ row of $\hat{U}$, and $\{c_i\}_{i=1}^n$ have exactly $r$ unique vectors.  Similarly, for the population eigenvectors $U$ (top $\cl$   eigenvectors of $\kp$ or $X_0$), we define the population centroids as $(Z\nu)_i$ , for some $\nu\in \bR^{\cl\times \cl}$.  The following theorem shows that as long as the number of outliers is not too large, then the inliers will be scattered in exactly $r$ clusters.

\begin{theorem}
Let $\hat{V} \in \bR^{n\times r}$ be the input eigenvectors of \km and $V$ be some eigenvectors of $n\times r$ such that $V$ has $r$ unique rows. Assume there exists rotation matrix $O$ such that $\|VO-\hat{V}\|\le u_{\hat{V}}$. If 
$3u^2_{\hat{V}}+2\frac{mr}{n} < 1$, then the inliers divided into exactly $r$ clusters.
\label{th:upper_bound_m}
\end{theorem}
The upper bound $u^2_{\hat{V}}$ can vary for different algorithms, and it is a function of $m$ and the eigengap of the population matrix. 
When we apply the upper bound generated from the Davis-Kahan Theorem, we can get some explicit sufficient condition for $m$, as stated in the following corollary.
\begin{corollary}
\begin{itemize}
\item[1.] Algorithm \ref{alg:sdp} returns exactly $r$ inlier clusters if $m < \frac{C_1n\gm}{r}$; 
\item[2.] Assume $\frac{p}{\log p}>2r+\frac{Cn^2}{(\lambda_r(\kp)-(\lambda_{r+1}(\kp))}$, then Algorithm \ref{alg:ksvd} returns exactly $r$ inlier clusters as long as  $m< \frac{C_2n}{\frac{n^2}{(\lambda_r-\lambda_{r+1})^2}+C'r} $. In particular, when all clusters share the same variance, Algorithm \ref{alg:ksvd} returns exactly $r$ inlier clusters if $ m< \frac{C_3n\gm^2}{r^2}$.
 \end{itemize}
\label{cor:upper_bound_m}
\end{corollary}

Theorem \ref{th:upper_bound_m} and Corollary \ref{cor:upper_bound_m} are proved in Appendix \ref{sec:proof_ub}.

We now show that when the empirical centroids are close to the population centroids with a rotation, then the node will be correctly clustered. 

We give a general definition of a superset of the misclustered nodes applicable both to \ksvd and SDP: 
\begin{align}
\cM=\{i: \|c_i-Z_i\nu O\|\ge 1/\sqrt{2n/\cl}\}
\label{eq:def_m}
\end{align}

\begin{theorem}
Let $\mathcal{M}_{sdp}$ and $\cM_{ksvd}$ be defined as Eq.~\ref{eq:def_m}, where $c_i$'s are generated from Algorithm \ref{alg:sdp} and \ref{alg:ksvd} respectively. Let $\lambda_\cl$ be the $\cl^{th}$ largest eigenvalue value of $\tilde{K}'$.
We have:
\begin{align*}
|\mathcal{M}_{sdp}| &\leq \max\left\{o_P(1), O_P\left( \frac{m}{\gm} \right)\right\} \\
|\mathcal{M}_{ksvd}| &\le O_P \max\left\{ \frac{mn^2}{\cl(\lambda_\cl-\lambda_{\cl+1})^2},\frac{n^3\log \dim}{\cl\dim(\lambda_\cl-\lambda_{\cl+1})^2} \right\}
\end{align*}
\label{th:mis_sdp}
\end{theorem}

\begin{remark}
Getting a bound for $\lambda_\cl$ in terms of $\gm$ for general blockwise constant matrices is difficult. But as shown in Lemma \ref{lem:eigengap_k}, the eigengap is $\Omega(n/\cl\lambda_{min}(B))$. Plugging this back in we have, 
\[ |\mathcal{M}_{ksvd}| \leq \max  \left\{O_P\left(\frac{m\cl}{\lambda_{min}(B)^2}\right),O_P\left(\frac{n\cl\log\dim/\dim}{\lambda_{\min}(B)^2}\right)\right\}\].
\end{remark}
In some simple cases one can get explicit bounds for $\lambda_\cl$, and we have the following.

\begin{corollary}
Consider the special case when all clusters share the same variance $\sigma^2$ and $d_{k\ell}$ are identical for all pairs of clusters. The number of misclustered nodes of \ksvd is upper bounded by:
\begin{align}
|\cM_{ksvd}| &\le \max \left(O_P\left( \frac{m\cl}{\gm^2} \right),O_P\left(\frac{n\cl\log\dim/\dim}{\gm^2}\right)\right)
\label{eq:mis_ksvd}
\end{align}
\label{cor:equal-dist}
\end{corollary}
Corollary \ref{cor:equal-dist} is proved in Appendix \ref{sec:proof_cor_ksvd}.

\begin{remark}
The situation may happen if cluster center for $a$ is of the form $c e_a$ where $e_a$ is a binary vector with $e_a(i)=\bm{1}_{a=i}$.
In this case, the algorithm is weakly consistent (fraction of misclassified nodes vanish) when $\gm = \Omega\left( \max\{ \sqrt{\frac{\cl\log\dim}{\dim}},\sqrt{\frac{m\cl }{n}} \} \right)$.
Compared to $|\cM_{sdp}|$, $|\cM_{ksvd}| $ an additional factor of $\frac{\cl}{\gm}$. With same $m,n$, the algorithm has worse upper bound of errors and is more sensitive to $\gamma_{\min}$, which depends both on the data distribution and the scale parameter of the kernel. The proposed SDP can be seen as a denoising procedure which enlarges the separation. It succeeds as long as the denoising is faithful, which requires much weaker assumptions.

\end{remark}

\section{Proof of the main results}
\label{sec:analysis}
In this section, we show the proof sketch of the main theorems. The full proofs are deferred to supplementary materials.

\subsection{Proof of Theorem~\ref{thm:kernel_tail}}

In Theorem~\ref{thm:kernel_tail}, we show that if the data distribution is subgaussian, the $\ell_\infty$ norm of $K-\tilde{K}$ concentrate with rate $O(\sqrt{\frac{\log \dim}{\dim}})$. 
\begin{proof}[Proof sketch]
With the Lipschitz condition, it suffices to show $\|Y_i-Y_j\|_2^2$ concentrates to $d_{k\ell}^2+\sigma_k^2+\sigma_\ell^2$. To do this, we decompose $\|Y_i-Y_j\|_2^2= \|\mu_k-\mu_\ell\|_2^2+2\frac{(W_i-W_j)^T}{\sqrt{\dim}}(\mu_k-\mu_\ell)+\frac{\|W_i-W_j\|_2^2}{\dim} $.
Now it suffices to show the third term concentrates to $\sigma_k^2+\sigma_\ell^2$ and the second term concentrates around 0. Note the fact that $W_i-W_j$ is sub-gaussian, its square is sub-exponential. With sub-gaussian tail bound and a Bernstein type inequality for sub-exponential random variables, we prove the result.
\end{proof}

With the elementwise bound, the Frobenius norm of the matrix difference is just one more factor of $n$.

\begin{corollary}
With probability at least $1-n^2p^{-\rho c^2}$, $\|K^{\cI\times \cI}-\kp^{\cI\times \cI}\|_F \le cn\sqrt{\log \dim /\dim}$.
\label{cor:k_fro}
\end{corollary}

\subsection{Proof of Theorem \ref{th:cst_sdp1}}
Lemma \ref{lem:ell1_dev} is proved in Appendix~\ref{sec:proof_lemma_ell1}, where we make use of the optimality condition and the constraints in SDP-1. Equipped with Lemma \ref{lem:ell1_dev} we're ready to prove Theorem \ref{th:cst_sdp1}.

\begin{proof}[Proof sketch]
In the outlier-free ideal scenario, Lemma \ref{lem:ell1_dev} along with the dualtiy of $\ell_1$ and $\ell_\infty$ norms we get $
\|\hat{X}-X_0\|_1 \le \frac{2\|K-\tilde{K}\|_\infty\|\hat{X}-X_0\|_1}{\gm }$. Then by Theorem~\ref{thm:kernel_tail}, we get the strong consistency result. When outliers are present, we have to derive a slightly different upper bound. The main idea is to divide the matrices into two parts, one corresponding to the rows and columns of inliers, and the other corresponding to those of the outliers. Now by the concentration result (Theorem~\ref{thm:kernel_tail}) on $K$ along with the fact that both the kernel function and $X_0, \xh$ are bounded by 1; and the rows of $\hat{X}$ sums to $n/\cl$ because of the constraint in SDP-1, we obtain the proof. The full proof is deferred to Appendix \ref{sec:proof_sdp1}.
\end{proof}

\subsection{Proof of Theorem \ref{th:mis_sdp}}
Although Theorem \ref{th:cst_sdp1} provides insights on how close the recovered matrix $\xhat$ is to the ground truth, it remains unclear how the final clustering result behaves. In this section, we bound the number of misclassified points by bounding the distance in eigenvectors of $\xhat$ and $X_0$. We start by presenting a lemma that provides a bound for \km step.

K-means is a non-convex procedure and is usually hard to analyze directly. However, when the centroids are well-separated, it is possible to come up with sufficient conditions for a node to be correctly clustered. When the set of misclustered nodes is defined as Eq.~\ref{eq:def_m}, the cardinality of $\cM$ is directly upper bounded by the distance between eigenvectors. To be explicit, we have the following lemma. Here $\hat{U}$ denotes top $\cl$ eigenvectors of $K$ for \ksvd and $\xh$ for SDP. $U$ denotes the top $\cl$ eigenvectors of $\kp'$ for \ksvd and $X_0$ for SDP. $O$ denotes the corresponding rotation that aligns the empirical eigenvectors to their population counterpart.
\begin{lemma}
$\cM$ is defined as Eq.~\eqref{eq:def_m}, then $|\cM| \le \frac{8n}{\cl} \|\hat{U}-UO\|_F^2$.
\label{lem:card_m}
\end{lemma}
Lemma \ref{lem:card_m} is proved in Appendix \ref{sec:proof_number_misclass}.

\textbf{Analysis of $|\cM_{sdp}|$:}
In order to get the deviation in eigenvectors, note the $\cl^{th}$ eigenvalue of $X_0$ is $n/\cl$, and $\cl+1^{th}$ is 0, let $U\in \bR^{n\times \cl}$ be top $\cl$ eigenvectors of $X$ and $\hat{U}$ be eigenvectors of $X_0$. By applying Davis-Kahan Theorem, we have
\begin{equation}
\begin{split}
\exists O, \|\hat{U}-UO\|_F  \leq \frac{2^{3/2}\|\hat{X}-X_0\|_F}{n/\cl}\leq \frac{\sqrt{8\|\hat{X}-X_0\|_1}}{n/\cl} =O_P\left( \sqrt{\frac{m\cl}{n\gm}}\right) \label{eq:eigen-x}
\end{split}
\end{equation}
Applying Lemma \ref{lem:card_m},
\begin{align*}
|\cM_{sdp}| \le& \frac{8n}{\cl} \left( \frac{2^{3/2}\|\xh-X_0\|_F}{n/\cl} \right)^2 \le \frac{cn}{\cl} \left( \sqrt{\frac{m\cl}{n\gm}} \right)^2 \le  O_P\left( \frac{m}{\gm} \right)
\end{align*}

\textbf{Analysis of $|\cM_{ksvd}|$:}
In the outlier-present kernel scenario, by Corollary~\ref{cor:k_fro}, 
\bas{
\|K-\kp'\|_F\leq \|K^{\cI\times \cI}-\kp^{\cI\times \cI}\|_F+\|K^{\cR}-\kp^{\cR}\|_F=O_P(n\sqrt{\log \dim/\dim })+O_P(\sqrt{mn})
}

Again by Davis-Kahan theorem, and the eigengap between $\lambda_\cl$ and $\lambda_{\cl+1}$ of $\kp$ from Lemma~\ref{lem:eigengap_k}, let $U$ be the matrix with rows as the top $\cl$ eigenvectors of $\kp$. Let $\hat{U}$ be its empirical counterpart.
\ba{
\label{eq:davis-k}
\exists O, \|\hat{U}-UO\|_F\leq \frac{2^{3/2}\|K-\kp\|_F}{\lambda_\cl-\lambda_{\cl+1}}\le O_P\left(\frac{\max\{\sqrt{mn},n \sqrt{\log \dim/\dim}\}}{\lambda_\cl-\lambda_{\cl+1}}\right)
}

Now we apply Lemma \ref{lem:card_m} and get the upper bound for number of misclustered nodes for \ksvd.
\begin{align*}
|\cM_{ksvd}| \le& \frac{8n}{\cl} \left( \frac{2^{3/2}  C\max\{\sqrt{mn},n \sqrt{\log \dim/\dim}\}}{\lambda_\cl(\kp)-\lambda_{\cl+1}(\kp)} \right)^2\\
\le& \frac{Cn}{\cl} \max\left\{ \left(\frac{\sqrt{mn}}{\lambda_\cl -\lambda_{\cl+1}} \right)^2, \frac{n^2\log\dim}{\dim(\lambda_\cl-\lambda_{\cl+1})} \right\} \\
\le&  O_P \max\left\{  \frac{mn^2}{\cl(\lambda_\cl-\lambda_{\cl+1})^2} ,\frac{n^3\log \dim}{r\dim(\lambda_\cl-\lambda_{\cl+1})^2} \right\}
\end{align*}

\section{Experiments}
\label{sec:exp}
In this section, we collect some numerical results. For implementation of the proposed SDP, we use Alternating Direction Method of Multipliers that is used in \cite{amini2014semidefinite}.
In each synthetic experiment, we generate $n-m$ inliers with $\cl$ equal-sized clusters. The centers of the clusters are sparse and hidden in a $p$-dim noise. For each generated data matrix, we add in $m$ observations of outliers. To capture the arbitrary nature of the outliers, we generate half the outliers by a random Gaussian with large variance (3 times of the signal), and the other half by a uniform distribution that scatters across all clusters.
We compare Algorithm \ref{alg:sdp} with 1) \km by Lloyd's algorithms; 2) kernel SVD and 3) kernel PCA by \cite{scholkopf1998nonlinear}. For all methods, we assume the number of clusters $\cl$ is known. In practice when dealing with outliers, it is natural to assume there is an extra cluster accounting for outliers, so we cluster both K-SVD and K-PCA with $\cl$ clusters and $\cl+1$ clusters.

The evaluating metrics are accuracy of inliers, i.e., number of correctly clustered nodes divided by the total number of inliers. To avoid the identification problem, we search for all permutations mapping predicted labels to ground truth labels and record the best accuracy. Each set of parameter is run 10 replicates and the mean accuracy and standard deviation (shown as error bars) are reported. For all \km used in the experiments we do 10 restarts and choose the one with largest objective. 
\begin{figure}[t]
\begin{tabular}{ccc}
\includegraphics[width=.30\linewidth]{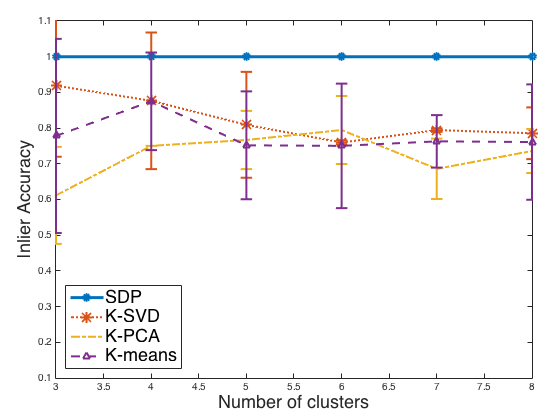}\hspace{-5in}&
\includegraphics[width=.30\linewidth]{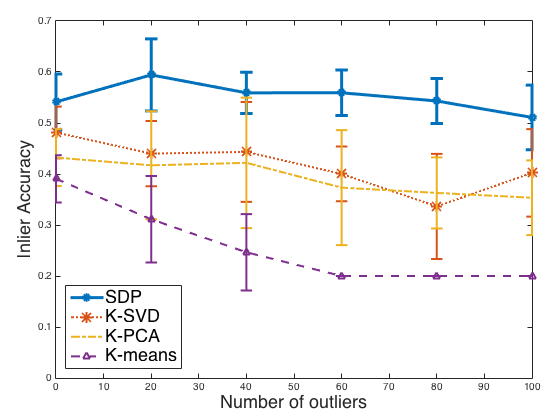} \hspace{-5in}& 
\includegraphics[width=.30\linewidth]{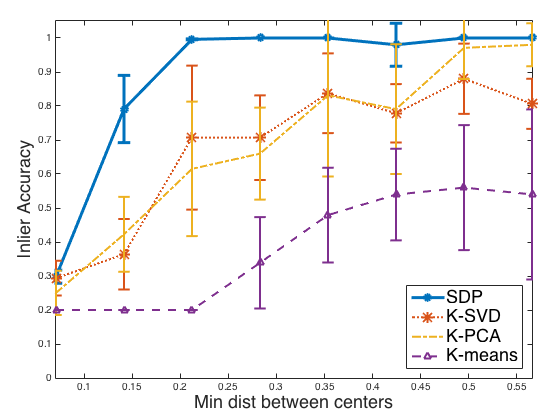} \hspace{-5em} \\
(a) \# clusters & (b) \# outliers & \quad  (c) Separation
\end{tabular}
\caption{Performance vs parameters: (a) Inlier accuracy vs number of cluster $(n=p=1500, m=10, d^2=0.125, \sigma=1)$; (b) Inlier accuracy vs number of outliers $(n=1000, \cl=5, d^2=0.02, \sigma=1, p=500)$; (c) Inlier accuracy vs separation $(n=1000, \cl=5, m=50, \sigma=1, p=1000)$.}
\label{fig:acc_p}
\end{figure}

For each experiment, we change only one parameter and fix all the others. Figure \ref{fig:acc_p} shows how the performance of different clustering algorithms change when (a) number of clusters (b) number of outliers (c) minimum distance between clusters increases. The value of all parameters used are specified in the caption of the figure. Setting number of clusters as $\cl+1$ doesn't help with clustering the inliers, which is observed in all experiments, the curve is then not shown here.

Panel (a) shows the inlier accuracy for various methods as we increase number of clusters. It can be seen that as we increase number of clusters in presence of outliers, the performance of all methods deteriorate except for the SDP, which matches the rate presented in Theorem \ref{th:mis_sdp}. We also examine the $\ell_1$ norm of $X_0-\xh$, which remains stable as the number of clusters increases. Note that the decrease in accuracy for \ksvd might result from the fact that it fails to meet the condition in Corollary \ref{cor:upper_bound_m}, which is stronger than the condition for SDP.
Panel (b) describes the trend with respect to number of outliers. The accuracy of SDP on inliers is almost unaffected by the number of outliers while other methods suffer with large $m$. Panel (c) compares the performance as the minimum distance between cluster centers changes. Both SDP and \ksvd are consistent as the distance increases. Compared to \ksvd, SDP concentrates faster and with smaller variation across random runs, which matches the analysis given in Section \ref{sec:mainres}.

\section{Conclusion}
In this paper, we investigate the consistency and robustness of two kernel-based clustering algorithms. We propose a semidefinite programming relaxation which is shown to be strongly consistent without outliers and weakly consistent in presence of arbitrary outliers. We also show that \ksvd is also weakly consistent in that the misclustering rate is going to zero as the observation grows and the outliers are of a small fraction of inliers. By comparing two methods, we conclude that although both are robust to outliers, the proposed SDP is less sensitive to the minimum separation between clusters. The experimental result also supports the theoretical analysis.

{\small
\bibliographystyle{plain}
\bibliography{Bibliography}}

%


\section*{Appendix}
\appendix
\section{Sub-gaussian random vector}
\label{sec:sub-gaussian}
In our analysis, we make use of some useful properties of sub-gaussian random variables, which are defined by the following equivalent properties. More discussions on this topic can be found in \cite{vershynin2010introduction}.
\begin{lemma}[\cite{vershynin2010introduction}]
The sub-gaussian norm of $X$ is denoted by $\|X\|_{\psi_2}$, 
\[ \|X\|_{\psi_2}=\sup_{p\ge 1} p^{-1/2}(\bE |X|^p)^{1/p}. \]
Every sub-gaussian random variable $X$ satisfies:
\begin{itemize}
\item[(1)] $P(|X|>t) \le \exp(1-ct^2/\|X\|^2_{\psi_2})$ for all $t\ge 0$;
\item[(2)] $(\bE |X|^p)^{1/p}\le \|X\|_{\psi_2}\sqrt{p}$ for all $p\ge 1$.
In particular, $\text{Var}(X)\le 2\|X\|_{\psi_2}^2 $.
\item[(3)] Consider a finite number of independent centered sub-gaussian random variables $X_i$. Then $\sum_i X_i$ is also a centered sub-gaussian random variable. Moreover,
\[ \|\sum_i X_i\|_{\psi_2}^2\le C\sum_i\|X_i\|_{\psi_2}^2 \]
\end{itemize}
\label{lem:subgaussian}
\end{lemma}
We say that a random vector $X\in \bR^n$ is sub-gaussian if the one-dimensional marginals $\innerprod{X}{x}$ are sub-gaussian random variables for all $x\in \bR^n$.

We will also see the square of sub-gaussian random variables, the following lemma shows it will be sub-exponential.
A random variable is sub-exponential if the following equivalent properties hold with parameters $K_i>0$ differing from each other by at most an absolute constant factor.
\begin{align}
P(|X|>t)\le \exp(1-t/K_1)\ \text{ for all } t\ge 0;\\
(\bE|X|)^{1/p} \le K_2p\ \text{ for all }p\ge 1;\\
\bE \exp(X/K_3)\le e.
\end{align}
\begin{lemma}[\cite{vershynin2010introduction}]
A random variable $X$ is sub-gaussian if and only if $X^2$ is sub-exponential. Moreover,
\[ \|X\|^2_{\psi_2}\le \|X^2\|_{\psi_1} \le2\|X\|_{\psi_2}^2 \]
\label{lem:subexp_sg2}
\end{lemma}

We have a Bernstein-type inequality for independent sum of sub-exponential random variables.
\begin{lemma}[\cite{vershynin2010introduction}]
Let $X_1,\cdots,X_N$ be independent centered sub-exponential random variable, and $M=\max_i\|X_i\|_{\psi_1}$. Then for every $a=(a_1,\cdots,a_N)\in \bR^{N}$ and every $t\ge 0$, we have
\[ P\left( \left| \sum_{i=1}^N a_iX_i\right|\ge t\right) \le 2\exp\left[-c\min\left( \frac{t^2}{M^2\|a\|_2^2},\frac{t}{M\|a\|_\infty} \right)\right] \]
where $c>0$ is an absolute constant.
\label{lem:subexp_tail}
\end{lemma}

\section{Proof of Theorem~\ref{thm:kernel_tail}}
To prove Theorem~\ref{thm:kernel_tail}, we work with the elementwise expansion, for ease of notation, we slightly abuse $K$ and $\tilde{K}$ to represent $K^{\cI\times \cI}$ and $\tilde{K}^{\cI\times \cI}$ in this proof. We use $c$ to represent any constant that does not depend on the parameters, and its value can change from line to line.
For $i\in C_k,j\in C_\ell$, recall that $W_i$ is sub-gaussian random vector with mean 0, covariance $\sigma_k^2I$ and sub-gaussian norm bounded by $\sgnorm$.
We have 
\begin{align}
\|Y_i-Y_j\|_2^2 = \|\mu_k-\mu_\ell\|_2^2+2\frac{(W_i-W_j)^T}{\sqrt{\dim}}(\mu_k-\mu_\ell)+\frac{\|W_i-W_j\|_2^2}{\dim}
\end{align}
As $W_i$ and $W_j$ are independent, $W_i-W_j$ has mean 0 and covariance $(\sigma_k^2+\sigma_\ell^2)I$.

Define
\begin{align*}
\beta_{ij}&=\|W_i-W_j\|_2^2/\dim-(\sigma_k^2+\sigma_\ell^2),  \\
\alpha_{ij}&=(W_i-W_j)'(\mu_k-\mu_\ell)/\sqrt{\dim}.\\
\end{align*}
Hence $\bE \beta_{ij}=0$. By the Lipschitz continuity of $f$, 
\begin{align}
 |K_{ij}-\tilde{K}_{ij}| &\le 2C_0 |\beta_{ij}+2\alpha_{ij}| 
\label{eq:kij}
\end{align}
By Lemma \ref{lem:subgaussian}-(3), $\alpha_{ij}$ is also sub-gaussian, with sub-gaussian norm upper bounded by
$2\sgnorm d_{k\ell}^2C/\dim$, for some $C>0$. Then by Lemma \ref{lem:subgaussian}-(1), $\exists C_1>0$ s.t.
\begin{align}
P\left(|\alpha_{ij}|\ge c\sqrt{\frac{\log \dim}{\dim}}\right)\le \dim^{-C_1c^2}
\end{align}



To bound $\beta_{ij}$, note each summand in Eq.~\eqref{eq:betaij_decomp} is a squared sub-gaussian random variable, 
thus is a sub-exponential random variable by Lemma \ref{lem:subexp_sg2}. 
\begin{align}
 \beta_{ij} = \sum_{d=1}^{\dim} (W_i^{(d)}-W_j^{(d)})^2/\dim -(\sigma_k^2+\sigma_\ell^2).
 \label{eq:betaij_decomp}
\end{align}

By Lemma \ref{lem:subexp_tail} with $t=c\sqrt{\frac{\log\dim}{\dim}}$, we see that with $a=(1,\dots,1)/p$, $\min\left( c^2\frac{t^2}{M^2\|a\|_2^2},c\frac{t}{M\|a\|_\infty} \right)=\min \left(\frac{c^2\log p}{M^2},\frac{c\sqrt{p\log p}}{M}\right)\geq c' \log p$ for large enough $p$. 
Thus $\exists C_2>0$ such that for large enough $p$,
\begin{align}
P\left(|\beta_{ij}| \le c\sqrt{\frac{\log \dim}{\dim}}\right) \ge 1-\dim^{-C_2c^2}
\label{eq:abs_beta}
\end{align}

By union bound, for some $\rho >0$, with probability at least $1-n^2\dim^{-\rho c^2}$,
\[\sup_{i,j\in \cI} |K_{ij}-\tilde{K}_{ij}|\le c\sqrt{\frac{\log \dim}{\dim}}.\]

\section{Proof of Lemma \ref{lem:eigengap_k}}
Define a diagonal matrix $D$ where $D_{ii}= f(\sigma_k^2), \text{ if } i\in C_k$ and 0 if $i\in \cO$. Write $\kp_0=\kp-I+D^2$, which is basically replacing the diagonal of $\kp$ to make it blockwise constant. 
By the fact $f(d^2_{k\ell}+\sigma_k^2+\sigma_\ell^2)=f(d_{k\ell}^2)f(\sigma_k^2)f(\sigma_\ell^2)$,
$\kp_0$ has the decomposition $\kp_0=DZBZ^TD$ where $B \in \bR_{\cl\times \cl}$ and $B_{k\ell}=f(d^2_{k\ell})$.
In fact, $B$ is exactly the Gaussian kernel matrix generated by $\{\mu_i\}_{i=1}^\cl$ centers, and is strictly positive semi-definite when the scale parameter $\eta \ne 0$ and centers are all different. Hence $\kp_0$ is rank $\cl$.
\begin{align*}
\lambda_\cl(DZBZ^TD)=\lambda_\cl(B^{1/2}Z^TD^2ZB^{1/2})=\lambda_\cl(BZ^TD^2Z)
\end{align*}

The first equality uses the fact that $XX^T$ and $X^TX$ has the same set of eigenvalues. The second step uses the fact that $B$ is full rank, since all clusters have distinct means. Now $B$ and $Z^TD^2Z$ are both $\cl\times \cl$ positive definite matrices. So the $\cl$th eigenvalue is the smallest eigenvalue. Now we use, $\lambda_{\min}(BZ^TD^2Z)\geq \lambda_{\min}(B)\lambda_{\min}(Z^TD^2Z)$ and have
\[ \lambda_{\cl}(\kp_0)\ge \lambda_\cl(Z^TD^2Z) \lambda_\cl(B)\ge \frac{n}{\cl}\lambda_{\min}(B)\cdot \min_k\left(f(\sigma_k^2)\right)^2.\]
Then $\lambda_\cl(\kp_0)=\Omega(\frac{n}{\cl})$.
On the other hand, $\|I-D^2\|_2\le \max_k (1-f(2\sigma_k^2))$. Let $\lambda_\cl(\kp),\lambda_{\cl+1}(\kp)$ be the $\cl^{th}$ and $\cl+1^{th}$ eigenvalue of $\kp$, by Weyl's inequality, 
\[ \lambda_\cl(\kp)\ge \lambda_\cl(\kp_0)-\max_k (1-f(2\sigma_k^2)) =\Omega(\frac{n}{\cl}\lambda_{\min}(B))\]
\begin{align}
 \lambda_{\cl+1}(\kp)\le \max_k (1-f(2\sigma_k^2)) =O(1) 
 \label{eq:lambda_kp1}
 \end{align}
Putting pieces together,
\[ \lambda_\cl(\kp)-\lambda_{\cl+1}(\kp)\ge  \frac{n}{\cl}\lambda_{\min}(B)\cdot \min_k\left(f(\sigma_k^2)\right)^2 - 2\max_k(1-f(2\sigma_k^2))=\Omega\left(\frac{n}{\cl}\lambda_{\min}(B)\right).\]

\section{Proof of Lemma~\ref{lem:ell1_dev}}
\label{sec:proof_lemma_ell1}

\begin{proof}
First note that $\xh$ is the optimal solution of \eqref{eq:sdp1}, so $\innerprod{K}{\xh}\ge\innerprod{K}{X_0}$. Hence $\innerprod{K-\kp}{\xh-X_0}\ge\innerprod{\kp}{X_0-\xh}$.

Let $a:=\min_k f(2\sigma_k^2)$, $b:=\max_{k\neq \ell} f(d_{k\ell}^2+\sigma_k^2+\sigma_\ell^2)$ and $\gamma_{min}:=a-b$, we have

\begin{equation}
\begin{split}
\innerprod{\kp}{X_0-\hat{X}} &= \sum_{k}\sum_{i\in \tC_k} \left( \sum_{j\in \tC_k} f(2\sigma_k^2)(1-\xh_{ij})- \sum_{\ell\neq k}\sum_{j\in \tC_\ell} f(d_{k\ell}^2+\sigma_k^2+\sigma_\ell^2)\xh_{ij} \right)  \\
&\geq \sum_{k}\sum_{i\in \tC_k} \left( a\sum_{j\in \tC_k} (1-\xh_{ij})-b\sum_{\ell\neq k}\sum_{j\in \tC_\ell} \xh_{ij}\right) \\
&\ge \sum_{k}\sum_{i\in \tC_k} \left( a\sum_{j\in \tC_k} (1-\xh_{ij})-b \left(\frac{n}{\cl}-\sum_{j\in \tC_k}\xh_{ij} \right) \right)  \\
&\ge \gm \sum_{k}\sum_{i\in \tC_k}\sum_{j\in \tC_k} (1-\xh_{ij})
\end{split} 
\label{eq:kx}
\end{equation}

On the other hand, by the fact that $\xh_{ij}\ge 0$ and row sum is $n/\cl$,
\begin{equation}
\begin{split}
\|X_0-\xh\|_1 &= \sum_k\sum_{i \in \tC_k} \left( \sum_{j\in \tC_k}(1-\xh_{ij})+\sum_{\ell\neq k}\sum_{j\in \tC_\ell}\xh_{ij}\right) \\
&= \sum_k\sum_{i\in \tC_k}\left( \sum_{j\in \tC_k}(1-\xh_{ij})+\left(n/\cl-\sum_{j\in \tC_k}\xh_{ij} \right) \right) \\
& \leq 2\sum_k\sum_{i\in \tC_k} \sum_{j\in \tC_k}(1-\xh_{ij}) 
\end{split}
\label{eq:xx}
\end{equation}

Equations~\eqref{eq:kx} and~\eqref{eq:xx} gives us:
\begin{align*}
\|X_0-\xh\|_1 \leq \frac{2}{\gm} \innerprod{\kp}{X_0-\hat{X}} \le\frac{ 2\innerprod{K-\kp}{\xhat-X_0} }{\gm}
\end{align*}
\end{proof}

\section{Proof of Theorem~\ref{th:cst_sdp1}}
\label{sec:proof_sdp1}
By Lemma \ref{lem:ell1_dev},
\begin{align*}
\|X_0-\xh\|_1 &\leq \frac{2 \innerprod{\kp}{X_0-\hat{X}}}{\gm}\le\frac{2 \innerprod{K-\kp}{\xhat-X_0}}{\gm}
\end{align*}
Divide the inner product into inlier part and outlier part, and note that $0<|K_{ij}-\kp_{ij}|<1, \forall i,j$. By Theorem \ref{thm:kernel_tail}, w.p. at least $1-n^2{\dim}^{-\rho c^2}$, we have
\begin{align*}
&\innerprod{K-\tilde{K}}{\hat{X}-X_0} \\
=& \innerprod{K^{\cI\times \cI}-\kp^{\cI\times \cI}}{\hat{X}-X_0}+\innerprod{K^\cR-\kp^\cR}{\hat{X}-X_0}\\
\le& \|\hat{X}-X_0\|_1\cdot \|K^{\cI\times \cI}-\tilde{K}^{\cI\times \cI}\|_\infty + \sum_{(i,j)\in \cR}(\xhat_{ij}-(X_0)_{ij})(K_{ij}-\kp_{ij}) \\
\le& \|\hat{X}-X_0\|_1\cdot \|K^{\cI\times \cI}-\tilde{K}^{\cI\times \cI}\|_\infty + \sum_{(i,j)\in \cR} \xhat_{ij}(K_{ij}-\kp_{ij}) - \sum_{(i,j)\in \cR}(X_0)_{ij}(K_{ij}-\kp_{ij}) \\
\le& \|\hat{X}-X_0\|_1\cdot \|K^{\cI\times \cI}-\tilde{K}^{\cI\times \cI}\|_\infty + \sum_{(i,j)\in \cR} \xhat_{ij} + \sum_{(i,j)\in \cR}(X_0)_{ij} \\
\le & C\sqrt{\frac{\log \dim}{\dim}}\|X_0-\hat{X}\|_1+\frac{4mn}{\cl}
\end{align*}

Thus,
\begin{align*}
(\gm-2C\sqrt{\frac{\log \dim}{\dim}})\|\hat{X}-X_0\|_1 \le \frac{4mn}{\cl}
\end{align*}

When $\sqrt{\frac{\log \dim}{\dim}} = o(\gamma_{\min})$, rearranging terms gives
\begin{align}
\|X_0-\xhat\|_1&\le \frac{\frac{4mn}{\cl}}{\gm-C\sqrt{\frac{\log p}{p}}} \\
&\le \frac{4mn}{\cl\gm} \left( 1+\frac{C}{\gm}\sqrt{\frac{\log \dim}{\dim}} \right) = O\left(\frac{mn}{\cl\gm}\right)
\end{align}

\section{Davis-Kahan Theorem}
\label{sec:proof_mis}
\begin{theorem}[\cite{yu2015useful}]
Let $\Sigma,\hat{\Sigma}\in\bR^{p\times p}$ be symmetric, with eigenvalues $\lambda_1\ge \cdots \ge \lambda_p$ and $\hat{\lambda}_1\ge \cdots \ge \hat{\lambda}_p$ respectively. Fix $1\le r\le s\le p$ and assume that $\min(\lambda_{r-1}-\lambda_r,\lambda_{s-1}-\lambda_s)>0$, where $\lambda_0:=\infty$ and $\lambda_{p+1}:=-\infty$. Let $d:=s-r+1$, and let $V=(v_r,v_{r+1},\cdots, v_s)\in \bR^{p\times d}$ and $\hat{V}=(\hat{v}_r,\hat{v}_{r+1},\cdots,\hat{v}_s)\in \bR^{p\times d}$ have orthonormal columns satisfying $\Sigma v_j=\lambda_j v_j$ and $\hat{\Sigma} \hat{v}_j=\hat{\lambda}_j \hat{v}_j$, for $j=r,r+1,\cdots,s$. Then there exists an orthogonal matrix $\hat{O}\in \bR^{d\times d}$ such that
\[ \|\hat{V}\hat{O}-V\|_F\le\frac{2^{3/2}\|\hat{\Sigma}-\Sigma\|_F}{\min(\lambda_{r-1}-\lambda_r,\lambda_{s-1}-\lambda_s)}. \]
\label{th:davis-kahan}
\end{theorem}

\section{Proof of Theorem \ref{th:upper_bound_m}}
\label{sec:proof_ub}
\begin{proof}
Let $R$ be a $n\times n$ matrix with $R(\cO,\cO)=I$ and zero otherwise, $\hat{V_\cl}=RV,\ \hat{V_\cO}=(I-R)V$. $\hat{V_\cI}^T\hat{V_\cI}=\hat{V}^T(I-R)\hat{V}$. For any input matrix $W$, define $\text{loss}_k(W) := \min_{M\text{ has exactly $k$ unique rows}} \|W-M\|_F^2$ as the \km loss of clustering $W$ corresponding to cluster number $k$.
Furthermore, define two feasible sets: $\mathcal{C}_1=\{ M\in \bR^{n\times \cl}: M_\cI \text{ has exactly }\cl \text{ unique rows} \}$ and $\mathcal{C}_2=\{ M\in \bR^{n\times \cl}: M_\cI \text{ has no more than }\cl-1 \text{ unique rows} \}$.
We want to obtain a condition such that 
\ba{
\min_{M\in \mathcal{C}_1} \|\hat{V}-M\|_F^2 < \min_{M\in \mathcal{C}_2} \|\hat{V}-M\|_F^2
\label{eq:km_condition}
}

Intuitively, this condition indicates the \km loss of inlier nodes assigned to no more than $r-1$ clusters is strictly larger than the \km loss for assigning inliers to exactly $r$ clusters. 
By optimality,
$\min_{M\in \cC_1}\|\hat{V}-M\|_F^2\leq \|\hat{V}-VO\|_F^2$, therefore a sufficient condition of Eq. \eqref{eq:km_condition} would be $\|\hat{V}-VO\|_F^2< \min_{M\in \mathcal{C}_2} \|\hat{V}-M\|_F^2$. Now, we will obtain a lower bound on the k-means loss on $\cC_2$. In order to do so, we will use \cite{overton1993optimality} to write the k-means loss for any number of clusters $k$ and input matrix $W$ as the following 0-1 SDP problem for any input matrix $W$. 
\bas{
\text{loss}_k(W)=	\min_{X}\quad & \tr(WW^T(I-X)), \\
s.t  \quad & X\bfone=\bfone, X=X^T,\  X\geq 0,\  \tr(X)=k, \ X^2=X.
}  

Note that by relaxing the constraints, we can see that:
\bas{
\text{loss}_k(W)\geq \min_{X} \tr(WW^T (I-X)),
	\quad \mbox{s.t}\ \    X=X^T, X^2=X, \tr(X)=k
}
The right hand side is essentially finding the trailing $k$ eigenvectors of $WW^T$ \cite{overton1993optimality}. Let the singular values of $W$ be $\sigma_1,\dots,\sigma_{\cl}$.

\ba{
 \text{loss}_k(W)\geq \sum_{i=k+1}^\cl\sigma_i^2 
 \label{eq:loss_k}
 }

Let $M^* = \arg\min_{M\in \cC_2} \|\hat{V}-M\|_F^2$, then 
\bas{
\min_{M\in \cC_2} \|\hat{V}-M\|_F^2 =& \|\hat{V}-M^*\|_F^2\\
=& \|\hat{V}_\cI-M^*_{\cI}\|_F^2 + \|\hat{V}_\cO-M^*_{\cO}\|_F^2 \\
\ge & \min_{s\le r-1} \text{loss}_s(\hat{V_\cI})+0
}
The last inequality comes from the fact that $M^*_{\cI}$ has no more than $r-1$ unique rows since $M^*\in \cC_2$.
Note that $\text{loss}_s$ is non-increasing as $s$ increases. To see this, consider the following procedure. 
Suppose the solution for $(k-1)$ centroids are $\{c_i\}_{i=1}^{k-1}$, now generate a feasible $k$ centroid solution by keeping $\{c_i\}_{i=1}^{k-1}$ and picking the $k^{th}$ centroid as the point that has largest distance with its corresponding centroid (there will always exist such a point that does not overlap with the existing centroids as long as loss is greater than 0). This consists an upper bound for the \km loss with $k$ clusters, which is smaller than the \km loss with $k-1$ clusters.  

Therefore without loss of generality, we assume the inliers are assigned $\cl-1$ clusters and one cluster contains only outliers. By Eq.~\eqref{eq:loss_k} we have
\bas{
 \text{loss}_{\cl-1}(\hat{V}_{\cI})\geq \sigma_\cl(\hat{V}_{\cI})^2=\lambda_\cl(\hat{V_\cI}^T\hat{V_\cI})\geq \lambda_\cl(\hat{V}^T\hat{V})-\|\hat{V}^TR\hat{V}\| \geq 1-\|\hat{V}_\cO\|_F^2
 }
Now, $\|\hat{V}_\cO\|_F\leq \|V_\cO O\|_F+\|\hat{V}_\cO-V_\cO O\|_F$
However, recall that $V=Z\nu$, and since $V^TV=I_\cl$, $\nu^T\nu=\cl/nI$. Thus every row of $V$ is of norm $\sqrt{\frac{\cl}{n}}$. Using $(a+b)^2\leq 2(a^2+b^2) $, we have:
\bas{
	\|\hat{V}_\cO\|_F^2\leq 2(\|V_\cO O\|^2_F+\|\hat{V}_\cO-V_\cO O\|^2_F)\leq 2\left(\frac{mr}{n}+\|\hat{V}-V O\|^2_F\right)
}

Let $u^2_{\hat{V}}$ denote an upper bound on $\|\hat{V}-V O\|^2_F$, then we have:
\bas{
\text{loss}_{\cl-1}(\hat{V}_{\cI})\geq 1-2\left(\frac{mr}{n}+u^2_{\hat{V}}\right)
}
On the other hand, $\text{loss}_{\cl}(\hat{V})\leq \|\hat{V}-VO\|_F^2\leq u^2_{\hat{V}}$ by optimality. Hence, we use the condition,

\begin{align}
1-2\left(\frac{mr}{n}+u^2_{\hat{V}}\right)\geq u^2_{\hat{V}}\qquad \Rightarrow \qquad 3u^2_{\hat{V}}+2\frac{mr}{n} < 1
\label{eq:u_hat_r_ineq}
\end{align}

\end{proof}

\paragraph{Proof of Corollary~\ref{cor:upper_bound_m}}
\begin{proof}
By Eq.~\eqref{eq:eigen-x}, we have for eigenvectors of $\xh$,
\bas{
\|\hat{U}-UO\|_F  \leq O_P\left( \sqrt{\frac{m\cl}{n\gm}}\right) 
}
Plug it to Theorem \ref{th:upper_bound_m} we have $u_{\hat{V}} = C\sqrt{\frac{mr}{n\gamma_{\min}}}$, therefore
\bas{
m <  \frac{n\gm}{r(C+2\gm)}=\frac{C'n\gm}{r}
}
For \ksvd, by Eq.~\eqref{eq:davis-k}, $u_{\hat{V}} = \max \left\{ O_P\left( \frac{\sqrt{mn}}{\lambda_r-\lambda_{r+1}} \right), O_P \left( \frac{n\sqrt{\log p/p}}{\lambda_r-\lambda_{r+1}}\right) \right\}$. 

We first consider the scenario where $m=O\left( \frac{n\log p}{p} \right)$, now $u_{\hat{V}} = \frac{C_1n\sqrt{\log p/p}}{\lambda_r-\lambda_{r+1}} $. Plugging this into inequality~\eqref{eq:u_hat_r_ineq}, we have 
\[ m<\frac{n}{2r}\left( 1-\frac{Cn^2\log p}{p(\lambda_r-\lambda_{r+1})^2} \right) \]

When $\frac{p}{\log p}>2r+\frac{Cn^2}{(\lambda_r(\kp)-(\lambda_{r+1}(\kp))^2}$, we have $\frac{n}{2r}\left( 1-\frac{Cn^2\log p}{p(\lambda_r-\lambda_{r+1})^2} \right) > \frac{n\log p}{p} $, therefore $m= O\left( \frac{n\log p}{p} \right) = O\left( \frac{n}{2r+\frac{Cn^2}{(\lambda_r-\lambda_{r+1})^2}} \right)$.

In the second scenario where $m=\Omega\left( \frac{n\log p}{p} \right)$, we have $u_{\hat{V}} =\frac{C_2\sqrt{mn}}{\lambda_r-\lambda_{r+1}}$. Now \eqref{eq:u_hat_r_ineq} solves
\begin{align}
m< \frac{Cn}{\frac{n^2}{(\lambda_r-\lambda_{r+1})^2}+C'r}
\label{eq:ubm_ksvd}
\end{align}
which shares the same formulation as the first condition.

In particular, when all clusters share the same variance, by Lemma~\ref{lem:eigengap_k}, $\lambda_r-\lambda_{r+1}=\Theta\left( \frac{n\gm}{r} \right)$. Substituting into Eq.~\eqref{eq:ubm_ksvd}, we have $m <  \frac{Cn\gm^2}{r^2} $.
\end{proof}

\section{Proof of Lemma \ref{lem:card_m}}
\label{sec:proof_number_misclass}
We prove the result for \km on $\xhat$. 
Let $\hat{U}$ be the top $\cl$ eigenvectors of $\xhat$, $U\in \bR^{n\times \cl}$ be the top $\cl$ eigenvector of $X_0$, then by construction, it can be written as $U=\begin{bmatrix} U^\cI \\ U^{\cO} \end{bmatrix} $. Let $\nu\in \bR^{\cl\times \cl}$ be the population value of the eigenvector corresponding to each cluster, $U=Z \nu$. 
$U$ is a unit basis so we know $I=U^TU=\nu^T Z^T Z \nu=\frac{n}{\cl}\nu^T\nu$. So $\nu^T\nu=\frac{\cl}{n}I_\cl$.

Define $\mathcal{C}=\{ M\in \bR^{n\times \cl}: M \text{ has no more than}\cl \text{ unique rows} \}$. Then minimizing the \km objective for $\hat{U}$ is equivalent to 
\[ \min_{\{m_1,\cdots,m_\cl\}\subset \bR^\cl} \sum_i \min_g \|\hat{u}_i-m_g\|_2^2 =\min_{M\in \mathcal{C}} \|\hat{U}-M\|_F^2 \]
So $C=[c_1,\cdots, c_{n}]=\arg\min_{M\in \mathcal{C}} \|\hat{U}-M\|_F^2$ and $\|C-\hat{U}\|\le\|Z\nu O-\hat{U}\|$. $c_i$ is the center assigned to point $i$ by running \km on $\hat{U}$.

When $i,j\in \cI, Z_i\ne Z_j$, 
\begin{align*}
\|Z_i\nu-Z_j\nu\| =& \|(Z_i-Z_j)\nu\| \ge \sqrt{2}\min_{x:\|x\|^2=1} \sqrt{x^T\nu^T\nu x}=\sqrt{\frac{2\cl}{n}}
\end{align*}
So 
\begin{align}
\|c_i-Z_j\nu O\| \ge \|Z_i\nu-Z_j\nu\| -\|c_i-Z_i\nu O\|\ge \sqrt{\frac{2\cl}{n}}-\sqrt{\frac{\cl}{2n}}=\sqrt{\frac{\cl}{2n}}
\end{align}
Therefore when $i,j\in \cI$ and $Z_i\ne Z_j$, $\|c_i-Z_i\nu O\|<\sqrt{\frac{\cl}{2n}}\Rightarrow \|c_i-Z_i\nu O\|_2<\|c_i-Z_j\nu O\|_2$, which means node $i$ is correctly clustered.

Now we bound the cardinality of $\cM$.
\begin{align*}
|\cM| &\le \frac{2n}{\cl} \sum_{i\in \cI} \| c_i-Z_i\nu O \|_F^2\\
& = \frac{2n}{\cl} \|C^\cI-U^\cI O\|_F^2\\
& \le \frac{2n}{\cl} (\|C^\cI-\hat{U}^\cI \|_F+\|\hat{U}^\cI-U^\cI O\|_F)^2\\
\|C^\cI-\hat{U}^\cI \|_F^2 &=  \|\hat{U}-C\|_F^2-\|C^\cO-\hat{U}^\cO\|_F^2\\
&\le  \|\hat{U}-C\|_F^2 \le \|\hat{U}-UO\|_F^2
\end{align*}
Therefore,
\[ |\cM| \le \frac{2n}{\cl} (\|\hat{U}-UO\|_F+\|\hat{U}^\cI-U^\cI O\|_F)^2 \le \frac{8n}{\cl} \|\hat{U}-U O\|_F^2\]

For \km procedure on $K$, note that $\kp'$ is blockwise constant except for the diagonals. It can be shown that the top $\cl$ eigenvectors of $\kp'$ are also piecewise constant. The rest of the analysis is similar to that of $\xh$.

\section{Proof of Corollary \ref{cor:equal-dist}}
\label{sec:proof_cor_ksvd}
\begin{proof}
Denote by $d_0$ the distance between clusters, $\alpha = f(2\sigma^2)$, $\beta=f(d_0^2+2\sigma^2)$, hence $\gamma_{\min}=\alpha-\beta$. Then $\kp$ has the form $(\alpha-\beta)X_0+\beta E+(1-\alpha)I$, and $\lambda_\cl(\kp)\geq \gamma_{\min} n/\cl$, since $\beta E+ (1-\alpha)I$ is positive semidefinite. 

On the other hand, from Lemma \ref{lem:eigengap_k} and Eq.~\eqref{eq:lambda_kp1}, $\lambda_{\cl+1}(\kp)\le 1-f(2\sigma^2)\le 1$. Hence $\lambda_\cl-\lambda_{\cl+1}\ge \frac{n}{\cl}\gamma_{\min}-1$. By Lemma \ref{lem:card_m} the misclassification rate of \ksvd becomes:
\begin{align*}
 |\cM_{ksvd}| &\le C\frac{n}{r}\left( \frac{2^{3/2}\|\kp-K\|_F}{\lambda_r(\kp)-\lambda_{r+1}(\kp)} \right)^2\\
 &\le C\frac{n}{r}\left( \frac{\max\left\{ n\sqrt{\frac{\log \dim}{\dim}},\sqrt{mn} \right\}}{\frac{n}{\cl}\gm} \right)^2 \\
& \le   \max \left(O_P\left( \frac{m\cl}{\gm^2} \right),O_P\left(\frac{n\cl\log\dim/\dim}{\gm^2}\right)\right)
\label{eq:mis_ksvd}
\end{align*}
\end{proof}

\end{document}